\documentclass[12pt]{article}

\usepackage[utf8]{inputenc}
\usepackage{amsmath}
\usepackage{amssymb}
\usepackage{physics}
\usepackage{mathtools}
\usepackage{comment}
\usepackage{amsthm}
\usepackage{bbm}
\usepackage{xcolor}
\usepackage{tikz-cd} 
\usepackage{adjustbox}
\usepackage[parfill]{parskip}
\usepackage{soul}
\usepackage[margin=1.375in]{geometry}
\newtheorem{theorem}{Theorem}[section]
\newtheorem{corollary}{Corollary}[theorem]

\newcommand{\set}[1]{\{ #1 \}}
\newcommand{\dprod}[2]{\langle #1, #2\rangle}

\begin{document}

\title{Memory and Capacity of Graph Embedding Methods}

\author{Frank Qiu \thanks{Statistics Department; University of California, Berkeley}}
        
\maketitle


\begin{abstract}
\textbf{THIS PAPER IS NOW DEFUNCT: Check out "Graph Embeddings via Tensor Products and Approximately Orthonormal Codes", where it has been combined into one paper.}
\end{abstract}


\section{Intro}
Previously, we introduced a graph embedding method that embeds the edge set of a graph \cite{Qiu_Recipe}. This method takes a "bind-and-sum" approach shared with other existing methods like \cite{Poduval_graph_embed} and  \cite{Nickel2016HolographicEO}. All such methods consist of two main components: a vertex code to represent vertices and a binding operation to generate edge embeddings. Our method uses spherical codes - vectors drawn uniformly from the unit hypersphere - and the tensor product as the binding operation. Other methods use alternative binding operations, such as the Hadamard product and convolution, along with their accompanying appropriate vertex codes. In this paper, we analyze and compare our binding-coding scheme to these other existing methods.

Our use of the tensor product as a binding operation is not new, and it was first proposed by Smolenksy \cite{Smolensky_tensor} for the general problem of role-filler bindings. However, in that context the tensor product had a major flaw: the size of the resulting tensor scaled poorly with both the number of objects bound together and the object dimension. Hence, various memory-efficient alternatives were proposed which preserved the object dimension, and at first glance it seems that these methods are more efficient in the number of parameters used. In this paper we shall investigate this claim in the context of graph embeddings. Specifically, we shall compare various coding/binding schemes to the spherical/tensor scheme on two aspects: the range of graph operations possible under each scheme as well as their memory-capacity ratio. With respect to the former, we show that these compressed alternatives are unable to represent some fundamental graph properties; for the latter, we find that any savings in the number of parameters is offset by a proportional penalty in the number of things able to be accurately stored in superposition. In other words, we find that they have the same memory-capacity ratio as the spherical/tensor scheme, while also sacrificing the ability to express fundamental graph properties.


\section{Notation and Terminology}
Our graph embedding method focuses on embedding objects - vertices, edges, graphs - into vector spaces. Therefore, we shall denote the objects using bolded letters and their embeddings with the corresponding unbolded letter. For example, we denote a graph using $\boldsymbol{G}$ and its embedding using $G$. In later sections, when there is no confusion we shall drop the distinction between the object and the embedding and use just the unbolded letter. We also adopt graph terminology that may be non-standard for some readers. There are many names for the two vertices of a directed edge $(d,c) = d \rightarrow c$.  In this paper, we shall call $d$ the domain of the edge and $c$ the codomain. 

\section{Spherical-Tensor Embedding Overview}
In  this section, we shall briefly introduce our graph embedding method, the spherical/tensor scheme.
Given some set $\boldsymbol{V}$, for any directed graph $\boldsymbol{G} = (\boldsymbol{V_G}, \boldsymbol{E_G})$ such that $\boldsymbol{V_G} \subseteq \boldsymbol{V}$ and $\boldsymbol{E_G} \subseteq \boldsymbol{V} \times \boldsymbol{V}$,  we embed $\boldsymbol{G}$ by embedding its edge-set $\boldsymbol{E_G}$ into a vector space. To this end, we first embed the large vertex set $\boldsymbol{V}$ by assigning each vertex to a $d$-dimensional spherical code - a vector drawn independently and uniformly from the unit hypersphere $\mathbb{S}^{d-1}$. We then embed each directed edge $(\boldsymbol{d},\boldsymbol{c})$ in the edgeset by the tensor product of their vertex embeddings:
\[
(\boldsymbol{d}, \boldsymbol{c}) \mapsto d \otimes c = d c^T
\]
Then, the embedding $G$ of the graph $\boldsymbol{G}$ - which we represent by its edgeset $\boldsymbol{E_G}$ - is the sum of its edge embeddings:
\[
\boldsymbol{G} \mapsto G = \sum_i d_i \otimes c_i = \sum_i d_i c_i^T
\]

\section{Alternative Binding Operations}
Our graph embedding method uses the tensor product to bind the vertex codes together into an edge. However, this means that the graph embedding space scales quadratically with the vertex code dimension, and in response various memory efficient alternatives have been proposed. In this and the subsequent section, we shall primarily analyze three prominent alternative binding operations: the Hadamard product, convolution, and circular correlation.

\subsection{Hadamard Product}
The Hadamard product of two vectors $a$ and $b$, denoted $a \odot b$, is their element-wise multiplication:
\[
[a \odot b]_k = a_k b_k
\]
It preserves the dimension of the vertex code and scales linearly with the code dimension. The most common vertex codes when using the Hadamard product are phasor codes and binary codes: vectors whose entries are the complex and real units respectively. These codes have nice unbinding operations with respect to the Hadamard product, where multiplication by the conjugate of a vertex code will remove that code from the binding:
\[
\overline{a} \odot (a \odot b) = b
\]
In the next section, we also briefly touch on other codes. However, we shall see that since unbinding the Hadamard product requires element-wise division, many choices of random codes are not suitable for accurate graph operations.

\subsection{Convolution}
The convolution of two vectors $a$ and $b$, denoted $a * b$, is defined as:
\[
[a * b]_k = \sum_i a_i b_{k-i} 
\]
If $\mathcal{F}$ denotes the Fourier transform, then the convolution can also be expressed as:
\[
a * b = \mathcal{F}^{-1}(\mathcal{F}(a) \odot \mathcal{F}(b) )
\]
Hence, convolution and the Hadamard product are equivalent up to a Hermitian change of basis, and so it is sufficient to analyze just the Hadamard product. Indeed, common codes used with convolution are those whose Fourier transforms are phasor and binary codes, and there is a bijection between the codes used for convolution and codes used for the Hadamard product.

\subsection{Circular Correlation}
The circular correlation of two vectors $a$ and $b$, denoted $a \star b$, is defined as:
\[
[a \star b]_k = \sum_i a_i b_{k+i}
\]
Using the Fourier transform, the circular correlation also can be expressed as:
\[
a \star b = \mathcal{F}^{-1}(\overline{\mathcal{F}(a)} \odot \mathcal{F}(b))
\]
Note that if $a$ is a real vector, then $\overline{\mathcal{F}(a)}$ is the Fourier transform of the flipped vector: let $y_k$ be the $k^{th}$ Fourier coefficient of $a$:
\[
y_k = \sum_{s=0}^{n-1} \exp(\frac{2\pi i (-ks)}{n}) a_s
\]
Taking the conjugate, we get:
\[
\overline{y_k} =  \sum_{s=0}^{n-1} \exp(\frac{2\pi i ( ks)}{n}) a_s =  \sum_{s'=0}^{n-1} \exp(\frac{2\pi i (-ks')}{n}) a_{-s'}
\]
Hence, $\mathcal{F}^{-1}(\overline{\mathcal{F}a} )$ = $Pa$, where $P$ is the permutation that flips indices, and so the circular correlation is convolution augmented with flipping the first argument. This is a special case of using permutation to induce ordered bindings, which we shall cover in the next subsection. Since the circular correlation is a permuted convolution and convolution is equivalent to the Hadamard product, it is sufficient to study the Hadamard product with permutations.

\subsection{Alternative Bindings as Compressions of the Tensor Product}
As noted in our derivation of the tensor product from superposition, the tensor product is a universal construction. That is, every binding method that respects superposition - bilinearity - is linearly induced from the tensor product. The three alternative binding operations - the Hadamard product, convolution, circular correlation- are no exception.\\
\\
More specifically, fixing the standard basis the tensor product $v \otimes w$ is the outer product $v w^T$, a $d \times d$ matrix. The Hadamard product is the main diagonal of this matrix. Similarly, the entries of the circular correlation are sums along pairs of diagonals of $v w^T$, spaced by an interval of $2(d-1)$.  For convolution, we analogously take sums along pairs of the reverse diagonals (bottom left to top right).\\
\\
Hence, all three operations can be seen as compressions of the tensor product, where we compress a $d \times d$ matrix into a $d$-dimensional vector. While on the surface this might seem to save much in terms of memory, in the next subsection we shall see these compressions are unable to perform some fundamental graph operations. Moreover, in the subsequent section we shall see that these binding alternatives do not actually save on memory, because they suffer from proportional decrease in their representational capacity.

\subsection{Hadamard Product: Graph Functionality}
As established in the preceding subsections, all three operations are, up to a Hermitian change of basis, equivalent to the Hadamard product with possibly a permutation applied to one of its arguments. Hence, in this section we shall analyze the Hadamard product and its suitability for graph embeddings.\\
\\
Firstly, the Hadamard product can perform edge composition when used in conjunction with binary codes. This is because it has easy unbinding operations, where multiplication by a code will remove that code from the binding. Using the binary/Hadamard scheme, we can perform edge composition by taking the Hadamard product of two edges:
\[
(a \odot b) \odot (b \odot c) = a \odot c
\]
Moreover, note that in the case of a mismatch between the vertices, the resulting edge is:
\[
(a \odot b) \odot (b' \odot c) = a \odot c \odot n
\]
where $n$ is a noisy binary code. Importantly, there is no destruction of mismatched edges during edge composition. This will impact its representation capacity, which we shall cover in the next section. Indeed, one can ask if there is any alternative edge composition function that can improve on using the Hadamard product. Assuming this edge composition function respects superposition, one can show that the natural edge composition function under the Hadamard product is again the Hadamard product. We give a sketch of the argument: firstly, the desired composition function respects superposition, so it is a multilinear function and is completely determined by its action on some basis $\set{b_i}$. Fixing the standard basis, we impose the natural constraint:
\[
(e_i \odot e_j) \times (e_k \odot e_l) \mapsto 
\begin{cases}
e_i \odot e_l & j = k\\
0 & j \neq k
\end{cases}
\]
Since these constraints are satisfied by the Hadamard product, our natural edge composition function must be the Hadamard product, and the defect of mismatched edges not interfering destructively during edge composition remains.\\
\\
Furthermore, the Hadamard product is unable to represent directed edges because it is symmetric: $a \odot b = b \odot a$. A common fix is to permute one of its arguments before binding, so now we bind as $Pa \odot b$ where $P$ in some permutation matrix. Then our augmented binding scheme becomes:
\[
(a,b) \mapsto Pa \odot b
\]
This augmented scheme is certainly able to represent directed edges, but it is now unable to perform edge composition. Indeed, disregarding oracle operations where one already knows the bound vertices, using the Hadamard product to perform edge composition results in:
\[
(Pa \odot b) \odot (Pb \odot c) = a \odot c \odot (b \odot Pb)
\]
The noise term $(b \odot Pb)$ will not cancel unless $P$ is the identity, the non-permuted case. A similar question arises whether there is a better edge composition funciton for the permutated Hadamard: assuming this edge composition function respects superposition, by multilinearity one can show no such operation exists. Intuitively, one would like to send the pair $(Pa,a)$ to the vector of all ones, but the symmetry in the Hadamard product and its compression means such a map is not possible without breaking multilinearity.\\
\\
In summary, we see that in the binary/Hadamard scheme one sacrifices some core graph functionality: with the regular Hadamard product, one can perform edge composition but cannot represent directed edges; in the permuted case, one can represent directed edges but cannot perform edge composition. Indeed, one can show that other core graph operations, like subsetting and graph homomorphisms, are also impossible under such schemes. Thus, in the binary/Hadamard case we see that compression hurts the representational power of the embedding method.

\subsubsection{Hadamard Product: Phasor Codes}
Phasor codes, which generalize binary codes, are even less suitable than binary codes. The natural unbinding operation for phasor codes is to multiply by the conjugate codes. However, this already makes it unsuitable for edge composition in both the non-permuted and permuted case. In the non-permuted case:
\[
\overline{(a \odot b)} \odot (b \odot c) = \Bar{a} \odot c \neq a \odot c 
\]
The permuted case has a similar deficiency. Again due to the compression of the Hadamard product, there is no multilinear function that conjugates just one argument of the bound edge $a \odot b$. Intuitively, this is due to the symmetry of the Hadamard product, since it unable to distinguish which particular vertex to conjugate. Edge composition using phasor codes is impossible, and any code which can be derived from the phasor code suffers a similar defect.

\subsection{Hadamard Product: Continuous Codes}
Similarly, since the unbinding the Hadamard product requires element-wise division, many choices of random continuous codes are numerically unstable. In fact, the next section we shall see that specific cases of continuous codes all suffer from having infinite moments, making accurate graph operations impossible since the noise terms will overwhelm the signal. Morever, operations like edge composition are also impossible for similar reasons as the phasor code, where one need to apply the unbinding operation to a specific vertex of the bound edge. Due the Hadamard product's compression, such a multilinear map does not exist.

\section{Random Codes} \label{Random Codes}
First, we state some facts about the different coding schemes under consideration. We shall assume all codes have common dimension $d$.

\subsection{Spherical Codes} \label{Sphere Codes}
We generate spherical codes by sampling iid from the $d$-dimensional unit hypersphere $\mathcal{S}^{d-1}$. They have the following properties:
\begin{theorem}[Spherical Code Properties]
Let $X$ denote the dot product between two spherical codes. Then, the following statements hold:
\begin{enumerate}
    \item $\frac{X+1}{2} \sim Beta(\frac{d-1}{2}, \frac{d-1}{2})$
    \item $E(X) = 0$ and $Var(X) = \frac{1}{d}$
    \item $|X| \propto \frac{1}{\sqrt{d}}$ with high probability.
\end{enumerate}
\begin{proof}
The first two claims follow from the results of \cite{Qiu_Recipe}. The final claim follows from either a standard Bernstein bound or a Gaussian approximation.
\end{proof}
\end{theorem}
\subsection{Rademacher Codes}\label{Rad Codes}
Rademacher codes are vectors $v$ where each entry is an iid Rademacher random variable: $v_i = \pm 1$ with probability $\frac{1}{2}$. They have the following properties.
\begin{theorem}[Rademacher Code Properties]
Let $X$ denote the dot product of two Rademacher vectors. The following statements hold:
\begin{enumerate}
    \item $\frac{X+d}{2} \sim Binom(d,\frac{1}{2})$
    \item $EX = 0$ and $Var(X) = d$
    \item $\frac{1}{d} X$ is approximately $N(0,\frac{1}{d})$.
    \item $X \propto \sqrt{d}$ with high probability.
\end{enumerate}
\begin{proof}
The first two claims follow from the fact that a sum of $n$ iid Rademacher variables is Binomial($n,\frac{1}{2}$).. The third claim follows either from the Central Limit Theorem or from the Gaussian approximation  to the binomial. The fourth claim follows from this same Gaussian approximation or a standard Chernoff inequality.
\end{proof}
\end{theorem}
Note that all Rademacher codes have norm $d$. Thus, we may scale them by $\frac{1}{\sqrt{d}}$ to normalize them, and in this sense they are special cases of spherical codes. In fact, by a similar argument as the spherical codes case, one can show that for a fixed error threshold of violating $\epsilon$-orthogonality, normalized Rademacher codes achieve the Johnson-Lindenstrauss upper bound.
However, one can have at most $2^d$ unique codes, while any finite number of spherical codes have probability zero of having a repeat. Rademacher codes still have a hard packing limit, but the trade-off is cleaner unbinding with respect to the Hadamard product.

\subsection{Other Continuous Codes}
Here, we will briefly describe three common continuous codes: Gaussian, Cauchy, and uniform codes. Gaussian codes are generated by having the component of each vector be drawn iid from some Gaussian - for now, let us assume the standard Gaussian. Cauchy and uniform codes are analogously generated - for now, let us assume the standard Cauchy and uniformly on the unit interval $[0,1]$. The main problem with these codes is that the Hadamard unbinding operation requires element-wise division:  the resulting ratios random variables will have infinite moments. This makes them unsuitable for accurate graph operations, since they will result in ill-controlled noise terms.
\begin{theorem}
For $t,u$ iid Gaussian, Cauchy, or uniform. Let $Y = \frac{t}{u}$. Then, for any of the three distributions, all moments of $Y$ are undefined.
\begin{proof}
In the Gaussian case, the ratio of two independent standard Gaussians is a Cauchy random variable, which is known to have infinite moments. In the Cauchy case, the ratio of two  independent standard Cauchy rv's has the density:
\[
f_{Y_c}(y) \propto \frac{1}{(y^2 - 1)}ln(y^2)
\]
Then, comparing integrals:
\[
\infty =\frac{1}{2} \int_0^{c}  |ln(y^2)| \leq \int_0^{c} y f_{Y_c} (y) \leq \int_0^\infty y f_{Y_c} (y)
\]
we see that the first moment is also undefined (for some sufficiently small constant $c$). In the uniform case, the ratio of two independent $U[0,1]$ rv's is:
\[
f_{Y_u}(y) = 
\begin{cases}
\frac{1}{2} & 0 < y < 1\\
\frac{1}{2z^2} & y \geq 1\\
0 & y \leq 0
\end{cases}
\]
A similar comparison test also shows that the first moment is undefined. Thus, the first moment, and hence all moments, are undefined for all three choices of distribution.
\end{proof}
\end{theorem}

\section{Binding Comparison Overview}
In a previous section, we showed that two other alternative binding methods - convolution and circular correlation - are special cases of the Hadamard product; similarly, we found that, of the codes considered, the binary code was the one that had the most representational power with respect to graph operations. Hence, we shall primarily analyze the memory and capacity of our graph embedding method relative to the Hadamard/Rademacher scheme. which uses random binary codes. We also briefly consider other continuous coding schemes paired with the Hadamard product, but we shall see that they are too noisy for accurate graph operations. 

We shall look at the memory vs. capacity tradeoff of different edge binding methods with respect to certain graph functions. Firstly, by the superposition principle we assume that the considered graph functions are multilinear functions. Secondly, we focus on two types of graph operations: first and second order operations. First order operations $f_1$ are any operations that involve graph embeddings once: $f_1(*,G)$ where $*$ represents non-graph arguments. Second order operations $f_2$ involve graph embeddings twice: $f_2(*,G,G')$. We analyze one representative graph operation from each type: vertex queries as a first-order operation and edge composition as a second-order operation. For both, we analyze the magnitude of the error term as well as the probability of error.

\section{Vertex Queries}
We denote the edge binding operation as $\psi$, which can be either the Hadamard product or the tensor product. Let us work with the following fixed graph:
\[
G = \psi(v,u) + \sum_{i=1}^{k} \psi(q_i,r_i)
\]
where all the $q,r$'s are distinct from $u,v$. We will perform an edge query that seeks to find the vertices in $G$ that vertex $v$ points to (or is connected  to in the undirected case).

\subsection{Hadamard Product and Rademacher Codes} \label{HR: Norms}
We first look at the Rademacher-Hadamard scheme, analyzing both the magnitude of the error term as well as bouding the probability of retrieving the correct vertex.
\subsubsection{Error Norms}
In this case, our vertex query is of the form:
\[
Q(v,G) = v \odot (v \odot u + \sum_{i=1}^k q_i \odot r_i) = u + \sum_{i=1}^k q_i \odot r_i \odot v
\]
Since the product of Rademcher's is still Rademacher, we see that each term $q_i \odot r_i \odot u = s_i$ is still a Rademacher random vector. We can express the output as:
\[
Q(v,G) = u + \sum_{i=1}^k s_i
\]
Thus, the result of our query can be split into the correct signal $u$ and a noise term $\epsilon$, which is a sum of $k$ independent Rademachers. We then have the following result on their expected magnitudes.
\begin{theorem}[Hadamard/Rademacher Signal-to-Noise]
When performing a vertex query with a single correct vertex $u$, under the Hadamard product and Rademacher codes of dimension $d$ we have:
\begin{enumerate}
    \item The squared norm of the signal $E||u||^2$ is $d$.
    \item The squared norm of the noise $E||\sum_{i=1}^k s_i||^2$ is $kd$
    \item The signal-to-noise ratio is $\frac{1}{k}$
\end{enumerate}
\begin{proof}
Let us consider the norm of the noise term $\sum_{i=1}^k r_i$ relative to the signal $u$. Then, using the results from Section $\ref{Rad Codes}$ and independence, we see that:
\[
E||u||^2 = d \qquad ; \qquad E||\sum_{i=1}^k r_i||^2 = \sum_i E||r_i||^2 + \sum_{j \neq k} E \dprod{r_j}{r_k} = kd
\]
Hence, their ratio is $\frac{1}{k}$
\end{proof}
\end{theorem}

Note that we assumed that all the edges were generated by independently sampling Rademacher codes, and hence precludes the possibility of a vertex participating in more than one edge. However, note that we can write the graph embedding into a sum of subgraphs such that each subgraph has edges whose vertices are all distinct:
\[
G = \sum_i^l G_i
\]
Note that the number of subgraphs is upper bounded by the maximum node connectivity of the graph $G$. That is, if every vertex in $G$ is connected to at most L other vertices, then it is possible to express $G$ as a sum of at most $2L$ subgraphs. Hence, we have the following corollary.

\begin{corollary}
If a graph $G$ has maximum connectivity $L$, then:
\begin{enumerate}
    \item The squared norm of the signal $E||u||^2$ is $d$.
    \item The squared norm of the noise $E||\sum_{i=1}^k s_i||^2$ is $4L^2kd$
    \item The signal-to-noise ratio is $\frac{1}{4L^2kd}$
\end{enumerate}
\end{corollary}
Note that this is a very loose result, since we count each of the $k$ edges $L$ times.

\subsubsection{Statistical Error}
Now, at this point our query function returns a superposition of the answer $u$ with a noise term $\sum_{i=1}^k s_i$. Hence, we would like to perform a look-up operation to recover $u$ by seeing which of the vertex embeddings the output $u + \sum_{i=1}^k s_i$ is most similar to. In this case, we will use the dot product to measure similarity. Due to the noise term, we might be concerned with the possibilities of recovering the wrong vertex. In particular, Theorem \ref{HR: Norms} suggests that as the number of edges increases, the probability of recovering an incorrect edge increases.\\
\\
First, consider the dot product of the true answer $u$ with the output:
\[
T = \dprod{u}{u + \sum_{i=1}^k s_i} = d + \sum_{i=1}^k \dprod{u}{s_i} = d + \epsilon
\]
Thus, the dot product of the true answer $T$ will have $ET = d$ and $Var(T) = kd$. The noise term  $\epsilon$ is a sum of $kd$ Rademacher random variables, and so $\frac{\epsilon+kd}{2} \sim Binom(kd,\frac{1}{2})$. The variance of $\epsilon$ is $kd$, and using the normal approximation to the binomial it is of order $\sqrt{kd}$ with high probability. Similarly, consider the dot product of a false vertex $v$ (that does not equal $s_i$ or $u$) with the output:
\[
F = \dprod{v}{u + \sum_{i=1}^k s_i} = \dprod{v}{u} + \sum_{i=1}^k \dprod{u}{s_i}
\]
Here, $F$ is the sum of $(k+1)d$ Rademacher random variables, with mean 0 and variance $(k+1)d$; moreover, $F$ is of order $\sqrt{(k+1)d}$ with high probability.\\
\\
Now, we  first approximate the probability that $F$ exceeds $d = ET$. We will use the CLT/Gaussian approximation to Binomial to approximate the sum of $n$ iid Rademachers $X$ as $N(0,n)$. Then, we can use these Gaussian bounds:
\[
\frac{C}{t} e^{-\frac{t^2}{2}} \leq P(X > t \sqrt{n} ) \leq  C e^{-\frac{t^2}{2}}
\]
where $C$ is some constant. Hence, applying this to $F$:
\[
C \sqrt{\frac{(k+1)}{d}} e^{-\frac{d}{2(k+1)}} \leq P(F > d) = P(F >  \sqrt{\frac{d}{(k+1)}} \sqrt{(k+1)d}) \leq e^{-\frac{d}{2(k+1)}}
\]
Thus, this suggests the limit of edges we can store in superposition and still have accurate recovery is $O(d)$. A similar computation shows the probability of $T$ being less than $0$ scales in a similar manner:
\[
C \sqrt{\frac{k}{d}} e^{-\frac{d}{2k}}  \leq P(T < 0) \leq e^{-\frac{d}{2k}}
\]
In fact, we can be a bit more precise and compute a lower bound on the probability of correct recovery.
\begin{theorem}
Under the above setup, let $A$ be the correct recovery event given $M$ erroneous choices: the event where the correct vertex $u$ is most similar to the output of the vertex query relative to $M$ other wrong candidate vertices. Then, for some constant $C$ we have:
\[
P(A) \geq 1 - M e^{-\frac{d}{2(2k+1)}}
\]
\begin{proof}
Now, let us first compute the lower bound. Note that correct recovery is precisely the event where the similarity of the correct vertex $T = \dprod{u}{u}$ is larger than the similarities $F_1\cdots, F_M$ of the $M$ erroneous vertices, where $F_i = \dprod{v_i}{u}$ for the erroneous vertex $v_i$. Then,
\[
P(T > \max(F_1,\cdots,F_M)) = P(\cap \set{T > F_i}) = 1 - P(\cup \set{T \leq F_i})
\]By construction, the $F_i$'s are iid. Letting $\epsilon$ denote  the error term:
\begin{align*}
  P(\cup \set{T \leq F_i}) &\leq \sum_{i=1}^M P(T \leq F_i)\\
  &= M P(T \leq F_1)\\
  &= M P(d + \epsilon \leq F_1)\\
  &= M P(F_1 - \epsilon \geq d)\\
  &\leq M e^{-\frac{d}{2(2k+1)}}
\end{align*}
Hence
\[
P(T > \max(F_1,\cdots,F_M)) \geq 1 - M e^{-\frac{d}{2(2k+1)}}
\]
We used the fact that a difference of Rademacher sums is still a Rademacher sum, so $F_1 - \epsilon$ is a sum of $(k+1)d + kd = (2k+1)d$ Rademachers. 
\end{proof}
\end{theorem}
This theorem confirms the informal analysis of this section: the number of edges $k$ that be stored in superposition cannot be more than $O(d)$ without seriously compromising the accuracy of the vertex query.

\subsection{Hadamard Product and Continuous Codes}
Now, let us suppose the we were working with any continuous code (Gaussian, Cauchy, Uniform). Our vertex query would now be unbinding the graph by the reciprocal of the query vertex $u$:
\[
Q(u,G) = u^{-1} \odot (u \odot v + \sum_{i=1}^k q_i \odot s_i) = v + \sum_{i=1}^k u^{-1} \odot q_i \odot s_i
\]
In the noise term, note that we now have a sum of vector whose entries are ratios: $\frac{q_i}{u^{-1}}s_i$. In section \ref{Random Codes}, we saw that the entries will have undefined moments: they follow heavy-tailed distribution. Thus, it is very likely that the noise overwhelms the true answer $v$ regardless of how many edges $k$ are in superposition. This makes such continuous codes infeasible for vertex queries.

\subsection{Tensor Product and Spherical Codes}
Here, we look at the same error quantities for the tensor-spherical scheme: the error norms and the probability of retrieving the correct vertex.
\subsubsection{Error Norms}
Now, our vertex query is of the form:
\[
Q(v, G) = v^T(vu^T + \sum_{i=1}^k q_i r_i^T) = u^T + \sum_{i=1}^k \dprod{v}{q_i}r_i^T = u^T + \sum_{i=1}^k s_i^T
\]
We have a corresponding result on the average squared norms and the signal-to-noise ratio.
\begin{theorem}[Tensor/Spherical Signal-to-Noise]
When performing a vertex query with a single correct vertex $u$, under the tensor product and spherical codes of dimension $d$ we have:
\begin{enumerate}
    \item The squared norm of the signal $E||u||^2$ is $1$.
    \item The squared norm of the noise $E||\sum_{i=1}^k s_i||^2$ is $\frac{k}{d}$
    \item The signal-to-noise ratio is $\frac{d}{k}$
\end{enumerate}
\begin{proof}
The first claim holds since spherical codes have norm 1. The squared norm of the nuisance term  $\sum_{i=1}^k \dprod{v}{q_i}s_i^T$ is:
\[
E||\sum_{i=1}^k \dprod{v}{q_i}s_i^T||^2 = \sum_i E(\dprod{v}{q_i})^2 + 2\sum_{j \neq k} E\dprod{v}{r_j}\dprod{v}{r_k}\dprod{r_j}{r_k} = \frac{k}{d}
\]
Hence, the ratio of the answer-noise average norms is $\frac{d}{k}$

\end{proof}
\end{theorem}

\subsubsection{Statistical Error}
Now, we again want to recover the answer $u$ by finding which vertex embedding the query output is most similar to, and we will again use the dot product to measure similarity. First, the dot product of the true answer $u$ with the query output:
\[
T = u^T (u + \sum_{i=1}^k \dprod{v}{q_i}r_i ) = 1 +  \sum_{i=1}^k \dprod{v}{q_i} \dprod{u}{r_i} = 1 + \epsilon
\]
The noise term $\epsilon$ is a sum of $k$ terms of the form $e_i = \dprod{v}{q_i} \dprod{u}{r_i}$, and using independence and the Cauchy-Schwarz inequality:
\[
E e_i = 0 \quad ; \quad E e_i^2 = \frac{1}{d^2} \quad ; \quad  E|e_i| \leq \frac{1}{d}
\]
Hence, the variance of $\epsilon$ is $\frac{k}{d^2}$, and so for accurate retrieval we see that $k \leq O(d^2)$ or else $\epsilon$ will be of the same magnitude as the signal 1. Similarly, the dot product of a false vertex $t$ (not matching any of the $v_i$'s) is:
\[
F = \dprod{t}{u} + \sum_{i=1}^k \dprod{v}{q_i} \dprod{t}{r_i} = \dprod{t}{u} + \epsilon
\]
Hence, let us first calculate the probabilty that $F$ exceeds the $ET = 1$. However, it is a sum of random variables with a different distribution, so we will make one further simplification. As in section \ref{Sphere Codes}, the term $ \dprod{t}{u}$ will be of the order $\frac{1}{\sqrt{d}}$ with high probability. Hence, we assume conservatively that $|\dprod{t}{u}| = O(\frac{1}{\sqrt{d}})$. Thus, for large $d$ we can make the following simplification:
\[
P(F > 1) \lessapprox P(\epsilon > 1 - O(\frac{1}{\sqrt{d}})) \approx P(\epsilon > 1)
\]
Hence, as $\epsilon = \sum^k e_i$ where $e_i$'s are independent with $Ee_i^2 = \frac{1}{d^2}$, then Bernstein's inequality gives:
\[
P(F > 1) \approx P(\epsilon > 1) \leq e^{-1/[2(\frac{k}{d^2} + \frac{1}{3})]} \approx e^{-\frac{d^2}{2k}}
\]
A similar computation. using the fact that $\epsilon$ is symmetrically distributed, gives an upper bound of $e^{-\frac{d^2}{k}}$ for $P(T < 0)$. Hence, both suggest that the limit of edges we can store in superposition an still have accurate recovery is $O(d^2)$.\\
\\
As in the Hadamard/Rademacher case, we have corresponding bound on the probability of accurate recovery for the tensor/spherical scheme.
\begin{theorem}
Under the Hadamard/Rademacher scheme, let $A$ be the correct recovery event given $M$ erroneous choices: the event where the correct vertex $u$ is most similar to the output of the vertex query relative to $M$ other wrong candidate vertices. Then, for some constant $C$ we have:
\[
P(A) \gtrapprox 1 - M e^{-\frac{d^2}{k}}
\]
\begin{proof}
First, we compute the lower bound.. We have:
\[
P(T > \max(F_1,\cdots,F_M)) = 1 - P(\cup \set{T \leq F_i})
\]
We can make the same simplifying conservative assumption of $|\dprod{t}{u}| = O(\frac{1}{\sqrt{d}})$ as above to get:
\begin{align*}
    P(\cup T \leq F_i) &\leq \sum P(T \leq F_i)\\
    &=  M P(T \leq F_1)\\
    &= M P(1 + \epsilon_1 \leq \dprod{t}{u} + \epsilon_2)\\
    &\lessapprox M P(\epsilon_2 - \epsilon_1 \geq 1 + O(\frac{1}{\sqrt{d}}))\\
    &\leq M e^{-\frac{d^2}{k}}
\end{align*}
Thus,
\[
P(T > \max(F_1,\cdots,F_M)) \gtrapprox 1 - M e^{-\frac{d^2}{k}}
\]
\end{proof}
\end{theorem}
Thus, this also confirms that when the vertex code dimension is $d$, we cannot store more than $d^2$ edges using the tensor/spherical scheme without compromising the accuracy of the vertex query.

\subsection{Memory and Capacity}
As a reminder, the vertex code has dimension $d$. Then, using the Hadamard product with Rademacher codes, the graph embedding space is also dimension $d$; the previous analysis suggests that we can store at most $k = O(d)$ edges in superposition without seriously affecting the accurate retrieval of the answer. On the other hand, using the tensor product with spherical codes, the graph embedding space is $d^2$, and the previous analysis shows that we can store at most $k = O(d^2)$ edge without affecting accuracy. Hence, in both cases the number of edges we can store in superposition vs. the dimension of the graph embedding space have the same ratio. 

\section{Edge Composition}
Now, let us work with the following fixed graph:
\[
G = \psi(u,v) + \psi(v,w) + \sum_{i=1}^{k-1} \psi(q_i,r_i)
\]
where all the $q,r$'s are distinct from $u,v,w$ and $\psi$ denotes the binding operation. We will look at edge composition, checking specifically for the correct composition of the two composable edges:
\[
(u,v) \circ (v,w) \mapsto (u,w)
\]
To check for the presence of the correct edge, we will perform an edge query and analyze both the error norms and probability of successfully retrieving the correct edge.
\subsection{Hadamard Product and Rademacher Codes}
In this section, we analyze edge composition in the Hadamard/Rademacher scheme.
\subsubsection{Error Norms}
We want to do edge composition with $G$, which in this case represents just the binding of $G$ with itself:
\[
G \odot G = (u \odot v + v \odot w + \sum_{i=1}^{k-1} q_i \odot r_i) \odot (u \odot v + v \odot w + \sum_{i=1}^{k-1} q_i \odot r_i)
\]
After distributing, we will get $(k+1)^2$ total terms:
\[
G \odot G = u \odot w + \sum_{i = 1}^{(k+1)^2 - 1} e_i = u \odot w + R
\]
We assumed that all every vertex was distinct, so each $e_i$ is a Hadamard product of either two or three vertices. Since a product of Rademachers is still Rademacher, each $e_i$ is a Rademacher vector and $\epsilon$ is a sum of $(k+1)^2 - 1 = k^2 + 2k$ independent Rademacher vectors.
\begin{theorem}[Hadamard/Rademacher Signal-to-Noise]
When performing a edge query with a single correct edge $(u,v)$, under the Hadamard product and Rademacher codes of dimension $d$ we have:
\begin{enumerate}
    \item The squared norm of the signal $E||u \odot w||^2$ is $d$.
    \item The squared norm of the noise $E||R||^2$ is $(k^2 + 2k)d$
    \item The signal-to-noise ratio is $\frac{1}{k^2}$
\end{enumerate}
\begin{proof}
By construction there is only one correct composable edge in $G$ - $u \odot w$ - and all other terms are noise. Using the results from section \ref{Rad Codes}, we can characterize the signal-to-noise ratio.
\[
E||u \odot w||^2 = d \qquad E||R|| = \sum_i E||e_i||^2 + \sum_{j \neq k}E \dprod{e_j}{e_k} = (k^2 + 2k)d
\]
\end{proof}
\end{theorem}

\subsubsection{Statistical Error}
After performing edge composition, we have a superposition of the single composed edge in $G$ - $u \odot w$ - along with noise $\epsilon$. Now, say we want to recover exactly which edges were composable in $G$. To this end, we can do two things: we can either unbind by one vertex and compute a dot product with the other, or we can unbind by the given edge and then sum all the entries together.  Both approaches give the same result, so we shall focus on the latter for simplicity.\\
\\
Hence, we shall detect the (non)existence of a candidate edge $s \odot t$ by first unbdinding and then summing the entries:
\[
G \odot G \mapsto (s \odot t) \odot (G \odot G) \mapsto sum[(s \odot t) \odot (G \odot G)]
\]
Now, let us first consider checking the true edge $u \odot w$:
\[
(u \odot w) \odot (u \odot w + E) = (1 + E')
\]
Unbinding by the true edge will generate a vector of 1's and, as a product of Rademachers is still Rademacher, a new error term $\epsilon'$ that, like $\epsilon$, is a sum of $k^2+k$ independent Rademacher vectors. Now, we then sum up the entries (or equivalently compute the dot product with the vectors of 1's) and we get:
\[
T = sum(1+E') = d + \sum^{d(k^2-2k)} r_i = d + \epsilon
\]
where each $r_i$ is a Rademacher random variable. Therefore, we use the same arguments as the vertex query section to get $E(T) = d$ and $Var(T) = d(k^2-2k)$. The noise term $\epsilon$ has variance $d(k^2-2k)$, and so it is of order $\sqrt{d(k^2-2k)} \approx k\sqrt{d}$. This suggests that for accurate retrieval, the number of edge in superposition $k \leq O(\sqrt{d})$.\\
\\
Similarly, we now do the same procedure for a false random edge $s \odot t$, and since it does not match then we will get  sum of $d(k^2-2k+1)$ Rademachers. Thus, the output $F$ of any false edge is:
\[
F = \sum^{d(k+1)^2} r_i
\]
We conclude that $EF = 0$ and $Var(EF) = d(k+1)^2$.\\
\\
Repeating the same analysis as in the vertex query section, we get the following bounds on the probability that $F$ exceeds $d = ET$:
\[
C \sqrt{\frac{(k+1)^2}{d}}e^{-\frac{d}{2(k+1)^2}}\leq P(F > d) \leq e^{-\frac{d}{2(k+1)^2}}
\]
Similarly, 
\[
C \sqrt{\frac{(k^2-2k)}{d}} e^{-\frac{d}{2k}}  \leq P(T < 0) \leq e^{-\frac{d}{2(k^2-2k)}}
\]
Finally, for $M$ false edges we have the following result using the same techniques as in the vertex query case:
\begin{theorem}
Under the Hadamard/Rademacher scheme, let $A$ be the correct recovery event given $M$ erroneous choices: the event where the correct edge $u \odot v$ is most similar to the output of the edge query relative to $M$ other wrong candidate edges. Then, for some constant $C$ we have:
\[ 
P(A) \geq 1 - M e^{-\frac{d}{4(k+1)^2-2}}
\]
\end{theorem}
Thus, these all suggest that for accurate edge composition, the number of edges in superposition can be at most $\sqrt{d}$.

\subsection{Tensor Product and Spherical Codes}
\subsubsection{Error Norm}
Using the tensor product, our graph $G$ is:
\[
G = uv^T + vw^T + \sum_{i=1}^{k-1} q_i r_i^T
\]
and we do edge composition by a matrix multiplication of the graph embedding:
\[
G^2 = uw^T + \sum_{i=1}^{(k+1)^2-1} \dprod{a_i}{b_i}d_i c_i^T = uw^T + R
\]
where $a,b,c,d$ are all iid uniform from the $d$-dimensional hypersphere.
\begin{theorem}[Tensor/Spherical Signal-to-Noise]
When performing a edge query with a single correct edge $(u,v)$, under the tensor product and spherical codes of dimension $d$ we have:
\begin{enumerate}
    \item The squared Frobenius norm of the signal $E||u w^T||_F^2$ is $1$.
    \item The squared Frobenius norm of the noise $E||R||_F^2$ is $\frac{k^2 - 2k}{d}$
    \item The signal-to-noise ratio is approximately $\frac{d}{k^2}$
\end{enumerate}
\begin{proof}
The signal, as the outer product of two orthonormal vectors, has Frobenius norm 1. Similarly, the nuisance term $E$ has the following expected squared Frobenius norm:
\[
||E(R^TR)||_F^2 = tr[E(R^TR)] = \sum_i E(\dprod{a_i,b_i})^2 = \frac{k^2 - 2k}{d}
\]
Hence, the ratio of the answer-noise average norms is $\frac{d}{k^2-2k} \approx \frac{d}{k^2}$.
\end{proof}
\end{theorem}
\subsubsection{Statistical Error}
Again, we query the edge composition $G^2$ for the (non)existence of a candidate edge. In the tensor product case, the natural edge query operation for a query edge $(s,t)$ is:
\[
s^T G^2 t
\]
Hence, let us first examine the result for the only true edge $(u,w)$:
\[
T = u^T(G^2)w = u^T (uw^T + R)w = 1 + u^T R w = 1 + \epsilon
\]
Expanding the error term $\epsilon$:
\[
\epsilon = u^T R w = u^T(\sum^{k^2-2k} \dprod{a_i}{b_i}d_ic_i^T) w = \sum^{k^2-2k}\dprod{a_i}{b_i} \dprod{u}{d_i} \dprod{c_i}{w}
\]
we see it is a product of $k^2 -2k$ iid terms, each of which is the product of three independent dot products. Hence, we have $E(\epsilon) = 0$ and $Var(\epsilon) = \frac{k^2-2k}{d^3} \approx \frac{k^2}{d^3}$, and similarly $ET = 1$ and $Var(T) = \frac{k^2-2k}{d^3}$. This suggests that $k \leq O(d^{\frac{3}{2}})$ to have accurate edge composition.\\
\\
Similarly, let us considering querying by any non-existent edge $s \rightarrow t$:
\[
F = s^T G^2 t = \dprod{s}{u}\dprod{w}{t} + \sum^{k^2-2k}\dprod{a_i}{b_i} \dprod{s}{d_i} \dprod{c_i}{t} =  \dprod{s}{u}\dprod{w}{t} + \epsilon
\]
The first term is a product of independent dot products, so it has mean $0$ and variance $\frac{1}{d^2}$. The second term has the exact same distribution as the error term in the previous paragraph. We have $EF = 0$ and $Var(F) = \frac{1}{d^2} + \frac{k^2 - 2k}{d^3}$\\
\\
Now, we first compute the probability that $F$ exceeds $ET = 1$. As in the edge binding section, using a Bernstein concentraation inequality the first term of $F$ - $\dprod{s}{u}\dprod{w}{t}$ - has magnitude at most $\frac{2}{d}$ with high probability. Hence, we work with the conservative assumption that $|\dprod{s}{u}\dprod{w}{t}| = O(\frac{1}{d})$. Hence, for large $d$ we can make the simplification:
\[
P(F > 1) \lessapprox P(\epsilon > 1 - O(\frac{1}{d})) \approx P(\epsilon > 1)
\]
Then, using a Bernstein inequality gives:
\[
P(F > 1) \lessapprox P(\epsilon > 1) \leq e^{-1/[2(\frac{k^2 + 2k)}{d^3} + \frac{1}{3})]} \approx e^{-\frac{d^3}{2k^2}}
\]
A similar computation for $T$, using the fact that $\epsilon$ is symmetrically distributed, gives:
\[
P(T < 0) \lessapprox  e^{-\frac{d^3}{2k^2}}
\]
These both suggest that we can store at most $O(d^{\frac{3}{2}})$ edges while retaining accurate recovery.

Finally, as in other sections we compute a lower bound on the probability of getting a correct answer when testing both the true edge $(t,w)$ and $M$ false edges.
\begin{theorem}
Under the tensor/spherical scheme, let $A$ be the correct recovery event given $M$ erroneous choices: the event where the correct edge $u v^T$ is most similar to the output of the edge query relative to $M$ other wrong candidate edges. Then, for some constant $C$ we have:
\[
P(A) \geq 1 -M e^{-\frac{d^3}{k^2}}
\]
\begin{proof}
Again, we make the same conservative assumption of $|\dprod{s}{u}\dprod{w}{t}| = O(\frac{1}{d})$.  We have:
\begin{align*}
    P(\cup \set{T \leq F_i}) &\leq \sum P(T \leq F_i)\\
    &= M P(T \leq F_i)\\
    &= M P(1 + \epsilon_1 \leq \dprod{s}{u}\dprod{w}{t} + \epsilon_2)\\
    &\lessapprox M P(\epsilon_2 - \epsilon_1 \geq 1)\\
    &\leq M e^{-\frac{d^3}{k^2}}
\end{align*}
Again, this suggests the number of edges $k$ in superposition must have order less than $d^{\frac{3}{2}}$ for accurate recovery.

\end{proof}
\end{theorem}

\subsection{Memory and Capacity}
The graph embedding dimension under Hadamard product with Rademacher codes is $d$, and the above analysis gives a limit of $\sqrt{d}$ edges that can be stored in superposition. On the other hand, the tensor product with spherical codes has dimension $d^2$ and it can store at most $d^{\frac{3}{2}}$ edges in superposition. In both cases, the edge-dimension ratio is $\sqrt{d}$. Again, the Hadamard product with Rademacher codes offers no concrete memory advantages over the tensor product. Any savings we have in memory are offset by a corresponding reduction in capacity.

\section{Binding Comparison Summary}

\subsection{General Memory vs. Capacity Ratio}
In general, for the Hadamard/Rademacher scheme a $n$-order operation on a graph with $k$ edges will create $k^n$ nuisance terms. Hence, a similar argument as the above two sections will give a limit of at most $d^\frac{1}{n}$ edges that can be stored in superposition. Thus, the general capacity-memory ratio for the Hadamard/Rademacher scheme is: 
\[
\frac{d^{1/n}}{d} = d^{\frac{1}{n} - 1}= d^{-\frac{n-1}{n}}
\]
On the other hand, for the tensor/spherical scheme an $n$-order operation will also create $k^n$ edges, but each will be weighted by a random coefficient with mean 0 and variance $d^{-(n-1)}$. During edge recovery, we will then have $k^n$ error terms with mean 0 and variance $d^{-(n+1)}$. Thus, our capacity-memory ratio for the tensor/spherical scheme is:
\[
\frac{d^{-(n+1)/n}}{d^2} = d^{-\frac{2n - (n+ 1)}{n}} = d^{-\frac{n-1}{n}} 
\]
In summary, we see that the Hadamard/Rademacher offers no relative memory advantages, since it suffers from  proportional hit to its capacity.

\subsubsection{Compression and Expressivity}
In the previous two sections, we saw that the Hadamard/Rademacher scheme not only falls short of the tensor/spherical scheme in terms of graph functionality, it also provides no meaningful savings in relative memory efficiency. Indeed, while we analyze just the specific Hadamard/Rademacher case, we can extend it to cover the other two alternative binding operations: these two alternatives are special cases of the Hadamard product, so we may as well analyze the Hadamard product and its possible codes; of the possible codes, the binary and phasor codes do not suffer from numerical and accuracy issues stemming from element-wise division as other codes do; of these two codes, the binary code is the only one capable of edge composition. Hence, from a graph functionality standpoint the Hadamard/Rademacher scheme is the closest to matching the tensor/spherical scheme, and so it is the natural scheme for comparison. However, we see that the Hadamard/Rademacher scheme still falls short in representational power and only matches in relative representational capacity of the tensor/spherical scheme.

A final point might be made about the effect of higher order tensors and their impact on memory, since the dimensionality of an $n$-order tensor is $d^n$. While this is general is a defect of the tensor product that the other alternative binding methods do not suffer from, in the specific context of graph embeddings the tensor order is always small; usually, it is at most 3 when working with typed graphs. Hence, concerns about memory stemming from higher order tensors do not apply to graph embeddings. In summary, from a memory standpoint the tensor product performs just as well as other alternative binding methods.

\section{Experimental Results}
In this section, we perform some simulations that confirm out theoretical results. For the spherical/tensor and Rademacher/Hadamard schemes, we looked at performance in the edge query and edge composition, comparing accuracy  in detecting both the presence and absence of a signal edge. In all cases, we looked at accuracy when varying the number of edges in superposition from 8 to 500; at each edge capacity, we generated a new graph and performed the one of the graph operations, repeating this for 200 trials and averaging the results. Each graph was generated by independently generating vertex codes, binding them, and summing them together. Shown below are the results of these experiments for the spherical/tensor case. The red lines denotes the ideal values: a value of 1 when the target edge was present and 0 when the target edge was absent.

\begin{figure}[!htb]
\begin{center}
\begin{tabular}{c  c}
\includegraphics[width=0.5\textwidth]{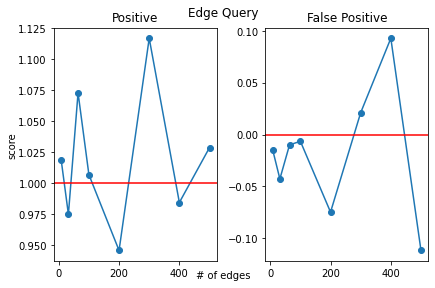}
    & \includegraphics[width=0.5\textwidth]{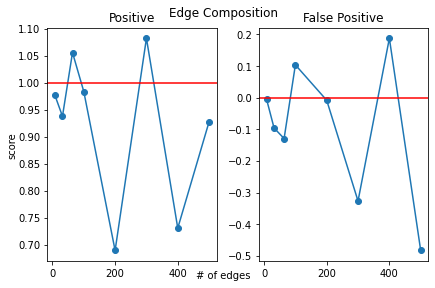}
\end{tabular}
\end{center}
\caption{Spherical/tensor scheme. Results are shown for the edge query and edge composition operations. For each graph operation, we tested performance in both detecting a test edge and ignoring a spurious edge. The red line indicate the ideal values in each case - 1 for the positive case and 0 for false positive case.}
\label{fig:1}
\end{figure}

Similarly, shown below are the same experiments for the Rademacher/Hademacher scheme. We see that the values deviate from the correct values much faster than the tensor spherical case.

\begin{figure}[!htb]
\begin{center}
\begin{tabular}{c  c}
\includegraphics[width=0.5\textwidth]{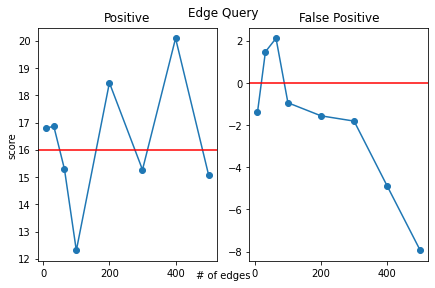}
    & \includegraphics[width=0.5\textwidth]{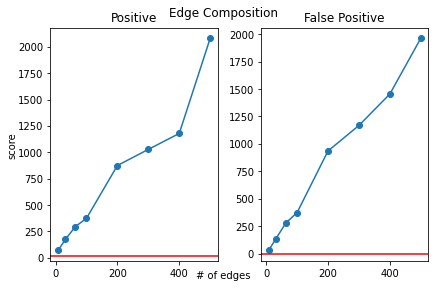}
\\
\end{tabular}
\end{center}
\caption{Rademacher/Hadmard scheme. We repeated the same tests from the tensor/spherical scheme. In this case, the ideal values for the positive and false positive cases are 16 and 0 respectively, denoted by red lines.}
\label{fig:2}
\end{figure}

We note that our theoretical analysis assumed that the graph embeddings had distinct edges, with no repeats. However, in these experiments we generated edges by binding together two vertex codes randomly sampled from a fixed codebook. While this codebook was randomly generated, this procedure does not exclude the possibility of vertex codes participating in multiple edges, which explains the monotonic behavior of the edge query/ edge composition tests. We chose this experimental setup because it is how graph embeddings are generated in practice, and the empirical results still corroborate our theoretical results: the capacity of the Rademacher/Hadamard scheme is much lower relative to the spherical/tensor scheme.


\section{Discussion}
Upon deeper analysis, we see that in the context of graph embeddings the purportedly memory-efficient Hadamard product offers no actual benefit over the tensor product, as any savings in memory is offset by a corresponding loss in capacity. Indeed, we saw that memory-capacity ratio of the Hadamard-Rademacher scheme merely matched that of the tensor-spherical scheme while also sacrificing some representational power. Furthermore, analyzing an idealized case of the vertex query showed that it is in fact impossible to compress the tensor product without adversely affecting accuracy. Hence, this suggests that any compressed binding operation suffers a similar penalty in the capacity like the Hadamard product and its relatives the convolution and circular correlation. This, combined with the generality of the tensor product \cite{Qiu_Recipe}, points to the tensor product as the natural candidate for edge binding in graph embeddings. Indeed, as the tensor product is a universal construction, many of the existing methods developed for other bind-and-sum graph methods, such as clean-up procedures, can be naturally adapted to the tensor product. In conclusion, we see that "memory-efficient" alternatives to the tensor product are not memory-efficient for graph embeddings. Moreover, our method can express many graph operations and properties that other methods cannot, and its flexibility allows it to adapt all multilinear procedures that were developed for other methods. In light of this, we argue that our method is a natural and convenient method for graph embeddings.

\clearpage

\bibliography{refs.bib}

\begin{thebibliography}{1}

\bibitem{Nickel2016HolographicEO}
Maximilian Nickel, Lorenzo Rosasco, and Tomaso~A. Poggio.
\newblock Holographic embeddings of knowledge graphs.
\newblock In {\em AAAI}, 2016.

\bibitem{Poduval_graph_embed}
Prathyush Poduval, Haleh Alimohamadi, Ali Zakeri, Farhad Imani, M.~Hassan
  Najafi, Tony Givargis, and Mohsen Imani.
\newblock Graphd: Graph-based hyperdimensional memorization for brain-like
  cognitive learning.
\newblock {\em Frontiers in Neuroscience}, 16, 2022.

\bibitem{Qiu_Recipe}
Frank Qiu.
\newblock Graph embeddings via tensor products and orthonormal codes.
\newblock 2022.

\bibitem{Smolensky_tensor}
Paul Smolensky.
\newblock Tensor product variable binding and the representation of symbolic
  structures in connectionist systems.
\newblock {\em Artificial Intelligence}, 46(1):159--216, 1990.

\end{thebibliography}
\bibliographystyle{plain}

\end{document}